\documentclass[11pt,a4paper]{article}
\usepackage[hyperref]{acl2021}
\usepackage{times}
\usepackage{latexsym}

\usepackage{microtype}

\usepackage{algorithm}
\usepackage{algorithmic}
\usepackage{mathrsfs}
\usepackage{amsmath}
\usepackage{graphicx}
\usepackage{booktabs}
\usepackage{array}
\usepackage{boldline}
\usepackage{multirow,tabularx}
\newcolumntype{C}[1]{>{\centering}m{#1}}

\aclfinalcopy 

\title{Annotation Inconsistency and Entity Bias in MultiWOZ}

\author{Kun Qian$^{\dagger}$\thanks{\hspace{0.15cm}The work of KQ and ZY was done as a research intern and a visiting research scientist at Facebook AI.}, 
Ahmad Beirami$^{\ddagger}$, Zhouhan Lin$^{\ddagger}$, Ankita De$^{\ddagger}$, Alborz Geramifard$^{\ddagger}$, \\
\textbf{Zhou Yu}$^{\dagger *}$\textbf{, Chinnadhurai Sankar}$^{\ddagger}$ \\
  $^{\dagger}$Columbia University \\
  \texttt{\{kq2157, zy2461\}@columbia.edu} \\
  $^{\ddagger}$Facebook AI\\
  \texttt{\{beirami, deankita, alborzg, chinnadhurai\}@fb.com}}
  
\date{}

\begin{document}
\maketitle
\begin{abstract}
MultiWOZ~\cite{budzianowski2018large} is one of the most popular multi-domain task-oriented dialog datasets, containing 10K+ annotated dialogs covering eight domains. 
It has been widely accepted as a benchmark for various dialog tasks, e.g., dialog state tracking (DST), natural language generation (NLG) and end-to-end (E2E) dialog modeling. 
In this work, we identify an overlooked issue with dialog state annotation inconsistencies in the dataset, where a slot type is tagged inconsistently across similar dialogs leading to confusion for DST modeling.
We propose an automated correction for this issue, which is present in 70\% of the dialogs. 
Additionally, we notice that there is significant entity bias in the dataset (e.g., ``cambridge" appears in 50\% of the destination cities in the train domain). The entity bias can potentially lead to named entity memorization in generative models, which may go unnoticed as the test set suffers from a similar entity bias as well. We release a new test set with all entities replaced with unseen entities.
Finally, we benchmark joint goal accuracy (JGA) of the state-of-the-art DST baselines on these modified versions of the data. Our experiments show that the annotation inconsistency corrections lead to 7-10\% improvement in JGA. On the other hand, we observe a 29\% drop in JGA when models are evaluated on the new test set with unseen entities. 
The data is available on the github.\footnote{\url{https://github.com/qbetterk/multiwoz/tree/master/data/MultiWOZ_2.2\%2B}}
\end{abstract}

\section{Introduction}
Commercial virtual assistants are used by millions via devices such as Amazon Alexa, Google Assistant, Apple Siri, and Facebook Portal. Modeling such conversations requires access to high quality and large task-oriented dialog datasets. 
Many researchers have devoted great efforts to creating such datasets and multiple task-oriented dialog datasets, e.g., WOZ~\cite{RojasBarahona2017ANE}, MultiWOZ~\cite{budzianowski2018large}, TaskMaster~\cite{byrne-etal-2019-taskmaster}, Schema-Guided Dialog~\cite{rastogi2019towards} with fine-grained dialog state annotation have been released in the recent years.

Among task-oriented dialog datasets, MultiWOZ~\cite{budzianowski2018large} has gained the most popularity. The dataset contains 10k+ dialogs and covers eight domains: \textit{Attraction, Bus, Hospital, Hotel, Restaurant, Taxi, Train} and {\it Police}. 
Each dialog can cover one or multiple domains. The inclusion of detailed annotations, e.g., task goal, dialog state, and dialog acts for both user side and system side, renders MultiWOZ a universal benchmark for many dialog tasks, such as dialog state tracking~\cite{Zhang2019FindOC, Zhang2020APE, Heck2020TripPyAT}, dialog policy optimization~\cite{Wu2019AlternatingRD, Wang2020ModellingHS, Wang2020MultiDomainDA} and end-to-end dialog modeling~\cite{Zhang2020TaskOrientedDS, HosseiniAsl2020ASL, Peng2020SOLOISTFT}.
Several recent papers, such as SimpleTOD~\cite{HosseiniAsl2020ASL}, TRADE~\cite{WuTradeDST2019}, MarCo~\cite{Wang2020MultiDomainDA}, evaluate their models solely on the MultiWOZ dataset, which makes their findings highly dependent on the quality of this dataset. 

\begin{figure*}
\begin{minipage}{0.71\textwidth}
 \includegraphics[width=\textwidth]{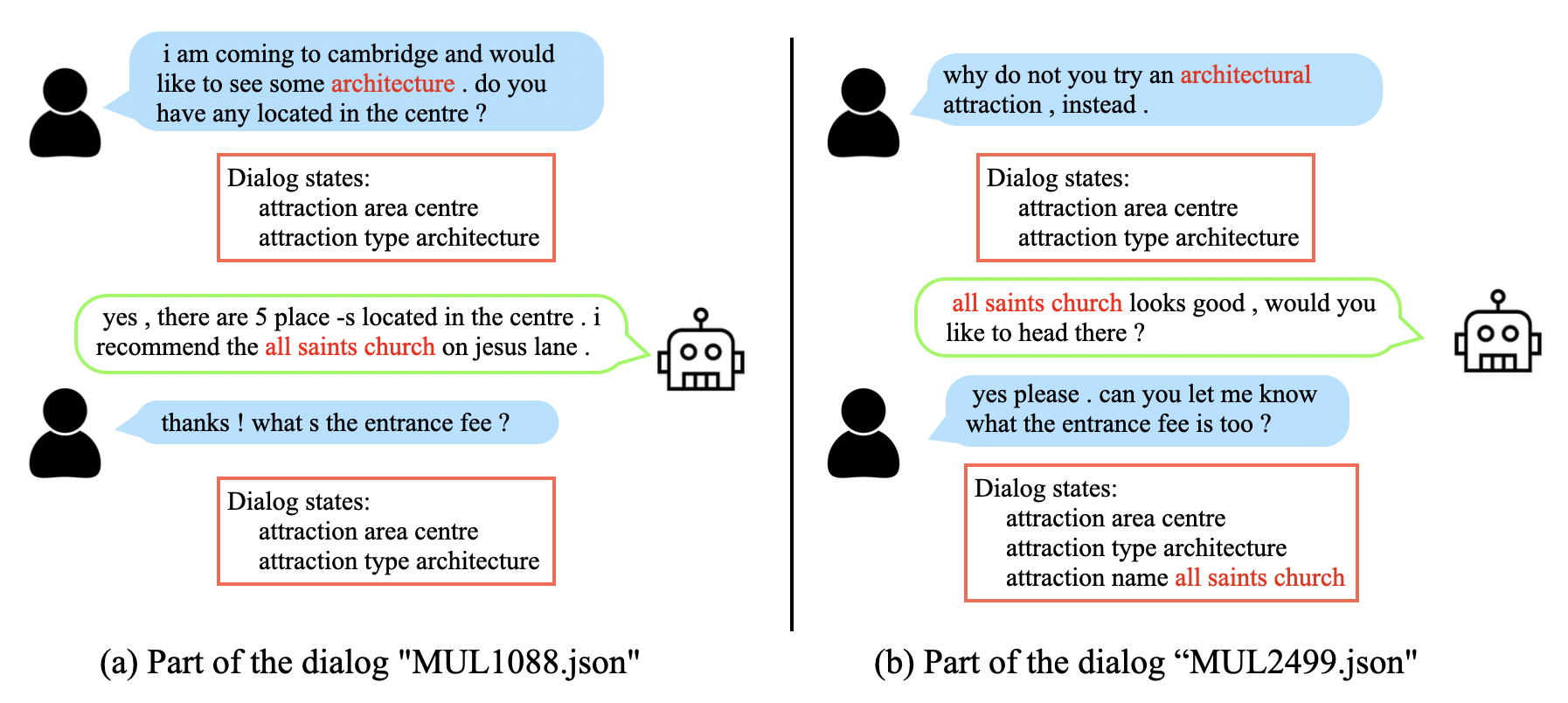}
 \caption{\small An example of two dialogs with inconsistent annotations. In the left dialog (a), the \textit{``attraction name"} mentioned in the system utterance is not annotated in the dialog state, while in the right dialog (b), the dialog state annotation includes the ``\textit{attraction name}".}
\label{fig_weights}
\end{minipage}
\hspace{0.02\textwidth}
\begin{minipage}{0.26\textwidth}
\scriptsize
\centering
\vspace{0.15in}
\begin{tabular}{lr}
\toprule
slot value & Count Num. \\
\hline
{\bf cambridge}               & {\bf 8,086}  \\
london liverpool street & 760   \\
leicester               & 746   \\
stansted airport        & 711   \\
stevenage               & 710   \\
ely                     & 695   \\
norwich                 & 692   \\
bishops stortford       & 667   \\
broxbourne              & 634   \\
peterborough            & 630   \\
birmingham new street   & 624   \\
london kings cross      & 609   \\
kings lynn              & 574   \\
\hline
total                   & 16,138 \\
\bottomrule
\end{tabular}
\vspace{0.13in}
\caption{\small The distribution of slot values for slot type ``destination" in the ``train" domain.}
\label{table_train_destination}
\end{minipage}
\vspace{-.15in}
\end{figure*}

Over the last couple of years, several sources of errors have been identified and corrected on MultiWOZ. 
\citet{WuTradeDST2019} pre-processed the dataset by normalizing the text and annotations. 
\citet{eric2019multiwoz} further corrected the dialog state annotations on over 40\% dialog turns and proposed MultiWOZ 2.1. 
Recently, \citet{zang2020multiwoz} identified some more error types, fixed annotations from nearly 30\% dialogs and added span annotations for both user and system utterances, leading to the MultiWOZ 2.2.  
Concurrent with this work, \citet{han2020multiwoz} released MultiWOZ 2.3 further looking at annotation consistency by exploring co-references.

While most of the previous works focus on correcting the annotation errors and inconsistencies within a dialog, where annotation contradicts the dialog context, we noticed another overlooked source of confusion for dialog state modeling, namely annotation inconsistency across different dialogs.
We first show that the dialogs have been annotated inconsistently with respect to the slot type `Name'. Figure~\ref{fig_weights} shows two dialogs in the {\it Attraction} domain with similar context.
In one dialog the attraction name is annotated while not in the other one. This inconsistency leads to a fundamental confusion for dialog state modeling whether to predict the attraction name or not in similar scenarios. In Section~\ref{sec:annotation_inconsistency}, we dive deeper into this problem and propose an automated correction for this problem.

We further found a second source of potential issue, entity bias, where the distribution of the slot value in the dataset is highly imbalanced. In Figure~\ref{table_train_destination}, we observed that ``cambridge" appears as the train destination in $50\%$ of the dialogs in train domain while there are $13$ destinations. 
As a result, a dialog system trained on this imbalanced data might be more likely to generate ``cambridge" as the slot value even though ``cambridge" might not even be mentioned in dialog history. In Section~\ref{sec:entity_bias}, we discuss this problem in more detail and suggest a new test set with all entities replaced with ones never seen during training. Finally, in Section~\ref{sec:experiments}, we benchmark the state-of-the-art dialog state tracking models on these new versions of data and conclude with our findings.

Our contributions in this paper can be summarized as follows:
\begin{itemize}
    \item We identify annotation inconsistency across similar dialogs as a new source of error that leads to confusion for DST modeling. We also propose an automated correction for these inconsistencies which result in changes in 66\% of the dialogs in the dataset, and release the new training/validation/test data.
    \item We identify that several slot types suffer from severe entity bias that potentially lead to models memorizing these entities, and release a new test set where all entities are replaced with ones not seen in training data.
    \item We benchmark state-of-the-art DST models on the new version of data, and observe a 7-10\% performance improvement in joint goal accuracy compared to MultiWOZ 2.2. For the data bias, we observe that models evaluated on the new test set with unseen entities suffer from a 29\% performance drop potentially caused by memorization of these entities.
\end{itemize}

\begin{figure*}[t]
\centering
\includegraphics[width=\textwidth]{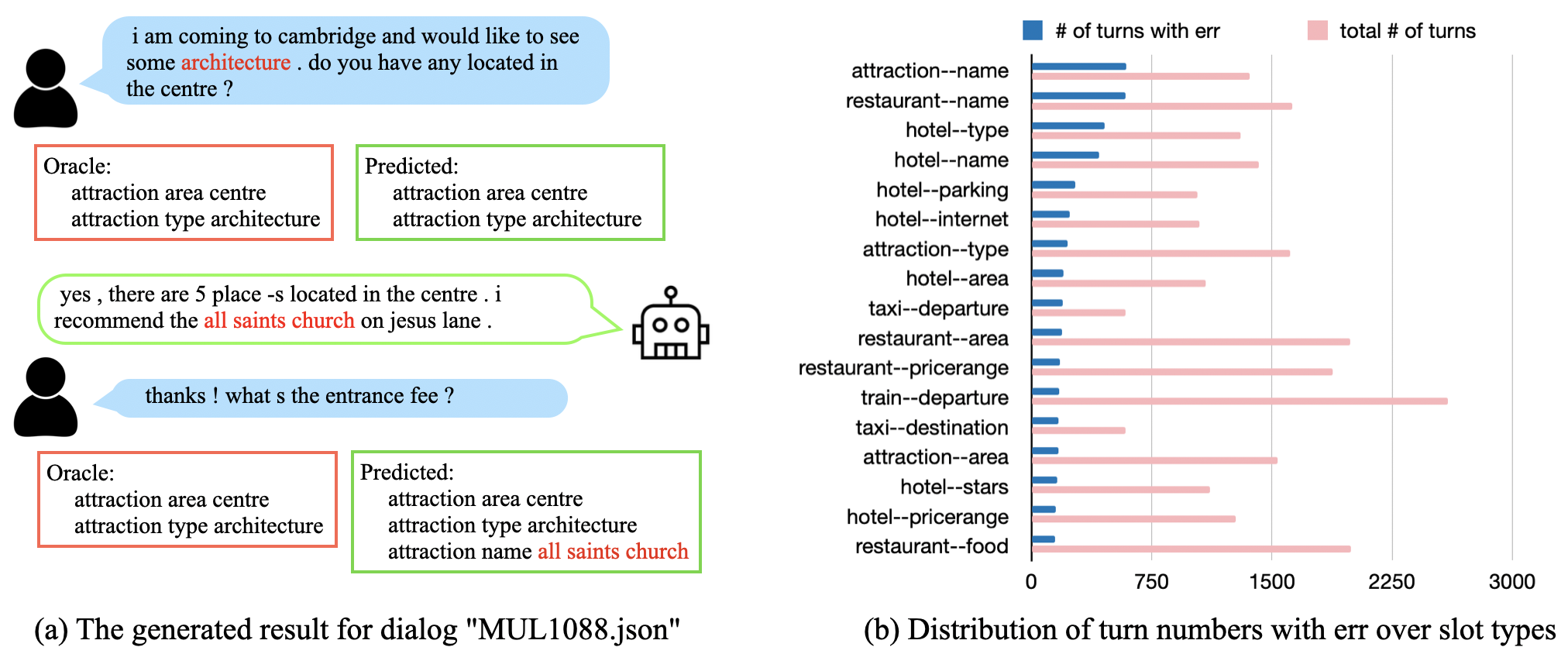}
\caption{(a) The generated dialog state results (in green rectangles) using SimpleTOD. The model generates the \textit{``attraction name all saints church"} in the second turn; (b) The distribution of the number of turns where SimpleTOD makes a mistake. The slot type ``\textit{name}" from domains ``\textit{attraction}", ``\textit{hotel}" and ``\textit{restaurant}" and slot type ``\textit{type}" from ``\textit{hotel}" domain have more error turns than others. 
}
\label{fig_inc_affect_model}
\vspace{-.1in}
\end{figure*}

\section{Related Work}
\label{multiwoz_diff}
\subsection{MultiWOZ 2.1}
MultiWOZ 2.1~\cite{eric2019multiwoz} mainly focuses on the semantic annotation errors. It identifies five main error types for dialog state annotation: delayed markups, multi-annotations, mis-annotations, typos and forgotten values. The delayed markups refer to the slot values that are annotated one or more turns after where the values show up in the utterances. Multi-annotations mean that multiple slots share the same slot value in a single turn. Mis-annotations represent the errors where a slot value is assigned to a wrong slot type, and the forgotten values refer to the missed annotations.

To solve those kinds of errors, \citet{eric2019multiwoz} adopted both manual corrections and automated corrections. After asking the human annotator to go over each dialog turn-by-turn, they also wrote scripts to canonicalize slot values to match the entities from the database. Besides, they also kept multiple slot values for over $250$ turns in the case that multiple values are included in the dialog context. In addition to correcting dialog states, MultiWOZ 2.1 also corrects typos within dialog utterances for better research exploration of copy-based dialog models. 
As a result, over $40\%$ of turns (around $30\%$ of dialog state annotations) are corrected.
Finally, MultiWOZ 2.1 also adds slot description for exploring few-shot learning and dialog act based on the pipeline from \cite{Lee2019ConvLabME}.

\subsection{MultiWOZ 2.2}
Building on version 2.1, MultiWOZ 2.2~\cite{zang2020multiwoz} further proposes four remaining error types for dialog state annotations: early markups, annotations from database, more typos and implicit time processing. Apart from those semantic errors, MultiWOZ 2.2 also identifies the inconsistency of the dialog state annotations. For example, a slot value can be copied from dialog utterance, or derived from another slot, or generated based on the ontology. They also identify issues with the ontology, e.g., the format of values is not consistent and $21\%$ of the values don't match the database.

MultiWOZ 2.2 designed a schema to replace the original ontology and divided all slots into two types: categorical and non-categorical. 
For categorical slots, the possible slot values are limited and less than $50$. 
Any value that is outside the scope of the database is labeled as ``unknown". 
On the other hand, values of non-categorical slots are directly copied from the dialog context and the slot span annotation is introduced to record the place and type of those non-categorical slots. 
Since typographical errors are inevitable in practice, MultiWOZ 2.2 leaves such errors in dialog utterances, hoping to train more robust models. 
In total, MultiWOZ 2.2 fixes around $17\%$ of dialog state annotations, involving around $30\%$ of dialogs.

In addition to the correction for the dialog states, MultiWOZ 2.2 also improve the annotations for the dialog acts. 
Though MultiWOZ 2.1 has added the dialog acts for the user side, there are still $5.82\%$ of turns ($8,333$ turns including both user and system sides) lacking dialog act annotations. 
After employing crowdsourcing to complete the annotations, MultiWOZ 2.2 also renames the dialog acts by removing the prefix, so that the annotation of dialog acts can be used across all domains. 

Our work builds on MultiWOZ 2.2, and further explores the annotation inconsistency (Section~\ref{sec:annotation_inconsistency}) and entity bias issues (Section~\ref{sec:entity_bias}) in the dataset.
%


\section{Annotation Inconsistency}
\label{sec:annotation_inconsistency}

MultiWOZ is collected following the Wizard-of-Oz setup~\cite{Kelley1984AnID}, where each dialog is conducted by two crowd-workers. One crowd-worker plays the role of a human user and another one plays the role of the dialog system.
The user would be assigned wh a task goal, which describes the information of the task object and is required to conduct the itconversation sticking to the task goal. 
The dialog system is required not only to respond to the user and complete the task, but also to take down the task-related information in the form of the slot values.
Since the slots is annotated by the crowd-worker, and different dialogs employ different crowd-workers, the strategies they decide whether to take the information down are also different.
Especially for the information provided by the dialog system, some crowd-workers decide to take it down while some do not.

\begin{figure}[t]
\centering
\includegraphics[width=0.97\columnwidth]{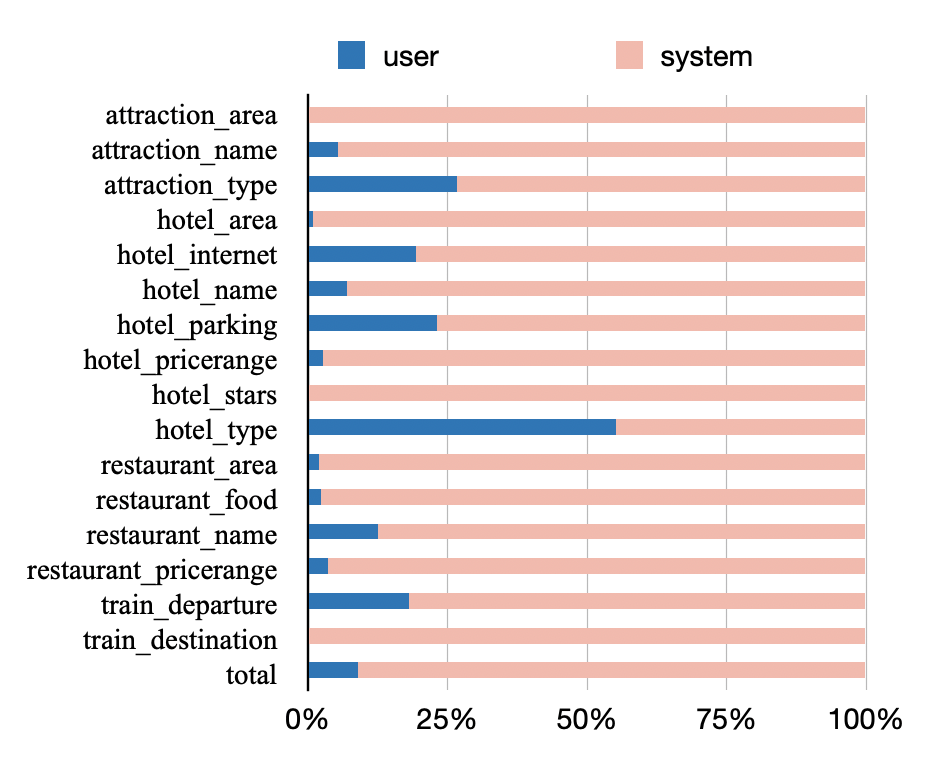}
\caption{\small The proportion of whether the newly-added slots are extracted from user utterances (dark blue) or system responses (light red). 
}
\label{fig_ratio}
\vspace{-.15in}
\end{figure}

In Figure~\ref{fig_weights}, we show two dialogs with similar context.
Both of the users ask for an \textbf{architectural attraction} in the \textbf{centre} area, and both of the dialog systems response with a result, \textbf{all saints church}. 
Then, both users acknowledge the result and ask for more information on the result.
However, in the left dialog, the attraction name, \textbf{all saints church}, is not annotated as one of the slot values, whereas the right dialog includes it in the dialog state.
The source of this discrepancy may be that the annotator in the left dialog thinks the dialog system already knows this information and only information provided by the user should be annotated. 
On the other hand, the annotator in the right dialog might notice that the user has acknowledged the result, attraction name, and asked questions based on this result, hence it should be included in the dialog state.
Having said that, the system cannot answer such follow-up questions without the annotation of the attraction name in the dialog state.
%
%
More examples of the inconsistency from other slot types or domains are listed in Figure~\ref{fig_more_examplt} in Appendix. 

\begin{table}[t]
\centering
\setlength{\extrarowheight}{0.06cm}
\scriptsize
\scalebox{0.95}{
\begin{tabular}[width=\textwidth]{c|l|lll}
\toprule
Domain                      & Slot Type   & Train         & Valid        & Test         \\\hline
\multirow{4}{*}{attraction} & area        & 491 (18.3\%)  & 86 (21.2\%)  & 76 (19.0\%)  \\
                            & name        & 1019 (38.0\%) & 151 (37.3\%) & 142 (35.6\%) \\
                            & type        & 674 (25.1\%)  & 102 (25.2\%) & 107 (26.8\%) \\\cline{2-5}
                            & total       & 1773 (66.1\%) & 283 (69.9\%) & 256 (64.2\%) \\\hline
\multirow{8}{*}{hotel}      & area        & 810 (24.0\%)  & 120 (28.6\%) & 97 (24.6\%)  \\
                            & internet    & 657 (19.5\%)  & 86 (20.5\%)  & 75 (19.0\%)  \\
                            & name        & 1319 (39.1\%) & 166 (39.6\%) & 150 (38.0\%) \\
                            & parking     & 638 (18.9\%)  & 87 (20.8\%)  & 71 (18.0\%)  \\
                            & pricerange  & 970 (28.8\%)  & 114 (27.2\%) & 104 (26.3\%) \\
                            & stars       & 665 (19.7\%)  & 106 (25.3\%) & 91 (23.0\%)  \\
                            & type        & 1460 (43.3\%) & 195 (46.5\%) & 185 (46.8\%) \\\cline{2-5}
                            & total       & 2907 (86.2\%) & 360 (85.9\%) & 346 (87.6\%) \\\hline
\multirow{5}{*}{restaurant} & area        & 799 (20.8\%)  & 99 (22.0\%)  & 82 (18.4\%)  \\
                            & food        & 689 (17.9\%)  & 80 (17.8\%)  & 88 (19.8\%)  \\
                            & name        & 1520 (39.6\%) & 189 (42.1\%) & 131 (29.4\%) \\
                            & pricerange  & 792 (20.6\%)  & 105 (23.4\%) & 82 (18.4\%)  \\\cline{2-5}
                            & total       & 2635 (68.6\%) & 318 (70.8\%) & 257 (57.8\%) \\\hline
\multirow{3}{*}{taxi}       & departure   & 18 (1.2\%)    & 1 (0.5\%)    & 0 (0.0\%)    \\
                            & destination & 14 (1.0\%)    & 1 (0.5\%)    & 1 (0.5\%)    \\\cline{2-5}
                            & total       & 31 (2.1\%)    & 2 (0.9\%)    & 1 (0.5\%)    \\\hline
\multirow{3}{*}{train}      & departure   & 70 (2.4\%)    & 5 (1.0\%)    & 8 (1.6\%)    \\
                            & destination & 124 (4.2\%)   & 19 (4.0\%)   & 1 (0.2\%)    \\\cline{2-5}
                            & total       & 185 (6.2\%)   & 23 (4.8\%)   & 8 (1.6\%)    \\\hline
\multicolumn{2}{c|}{total}                 & 5950 (74.2\%) & 768 (76.8\%) & 715 (71.5\%)\\
\bottomrule
\end{tabular}}
\caption{\small Number and (percentage) of modified dialogs to correct dialog annotation inconsistencies. }
\label{table_correct_num}
\vspace{-.15in}
\end{table}

After exploring all the dialogs from the test set, we manually examined and corrected each test dialog turn by adding the missing annotations.
%
%
We first checked whether there are any missing annotations from the user utterance. 
For example, in the sixth turn of dialog ``MUL0690.json", user is asking for ``\textit{a moderate hotel with free wifi and parking}". 
However, the token ``\textit{moderate}" from ``\textit{pricerange}" type is not included in the annotations.
So, we added the slot ``\textit{hotel pricerange moderate}" to this turn.
%
On the system side, for each dialog turn, we identified a possible slot value in the system response which is not included in the annotations, e.g., ``\textit{all saints church}" in the left dialog (``MUL1088.json) in  Figure~\ref{fig_weights}, which is determined to be a possible slot value from ``\textit{attraction}" domain and slot type ``\textit{name}" based on the database and ontology file.
Then, we examined those dialogs with slot annotations of ``\textit{attraction}" domain and slot type ``\textit{name}". 
If we could find such dialogs with similar dialog context and containing annotations of the same domain and slot type, 
we complemented the annotations by adding the missing slot value.

In Figure~\ref{fig_ratio}, we illustrate the fraction of added annotations that come from the user utterance vs. system side. Each row corresponds to a slot type from a certain domain. 
As can be seen, the majority of the added utterances are from the system side, which confirms our original hypothesis: annotators often have no disagreement to take down information from the user utterance. However, they have different opinions about whether to annotate slots based on system responses. 

\begin{table}[]
\footnotesize
\centering
\scalebox{1}{
\begin{tabular}[width=\columnwidth]{l|rrr}
\toprule
Domain--Slot$\_$Type      & $H_1/H_0$ & $H_{\infty}/H_0$ \\\hline
hotel--parking         & 0.217 & 0.060   \\
hotel--internet        & 0.225 & 0.053   \\
hotel--stars           & 0.592 & 0.249   \\
restaurant--food       & 0.638 & 0.377   \\
hotel--name            & 0.743 & 0.472   \\
train--destination     & 0.753 & 0.269   \\
hotel--stay            & 0.757 & 0.673   \\
train--departure       & 0.776 & 0.288   \\
attraction--area       & 0.792 & 0.355   \\
train--leaveat         & 0.801 & 0.681   \\
restaurant--area       & 0.824 & 0.384   \\
restaurant--time       & 0.833 & 0.758   \\
train--arriveby        & 0.850 & 0.732   \\
attraction--type       & 0.852 & 0.514   \\
attraction--name       & 0.855 & 0.636   \\
restaurant--name       & 0.877 & 0.709   \\
train--people          & 0.886 & 0.615  \\
taxi--arriveby         & 0.890 & 0.736   \\
taxi--departure        & 0.901 & 0.579   \\
hospital--department   & 0.926 & 0.530   \\
taxi--destination      & 0.936 & 0.685   \\
taxi--leaveat          & 0.942 & 0.781   \\
hotel--pricerange      & 0.944 & 0.662   \\
hotel--area            & 0.954 & 0.658   \\
hotel--type            & 0.969 & 0.729   \\
restaurant--pricerange & 0.971 & 0.746   \\
train--day             & 0.999 & 0.947   \\
hotel--day             & 0.999 & 0.954   \\
restaurant--day        & 0.999 & 0.955   \\
restaurant--people     & 0.999 & 0.969   \\
hotel--people          & 0.999 & 0.973   \\
\bottomrule
\end{tabular}}
\caption{\small The unbalanced distribution among different slot types, measured using $H_1/H_0$ (normalized Shannon entropy) and $H_{\infty}/H_0$ (normalized min-entropy).}
\label{table_bias}
\vspace{-.15in}
\end{table}

For the training and validation sets, we write regular expressions that match the test set corrections and apply the scripts to automatically correct the annotations based on the database and ontology file, and modify the dialogs automatically. 
Table~\ref{table_correct_num} list the corrected dialog numbers of each slot type and each domain, as well as the percentage of the corrected dialogs from all the dialogs in that domain. 
On average, about 20\% of the dialogs involved slot modification for each slot type. The ``name" and ``type" slot types involve the most modification with around 40\%.
As we mainly focus on the missing slots extracted from system responses in these automated scripts, we ignore the slot types that can be solely modified by the user utterance, such as ``book day" and ``book people". As can be seen, this process resulted in modification of totally more than $70\%$ of the dialogs .
%
To verify the correctness of the automated correction method, we randomly sampled $100$ modified dialogs and $100$ unchanged dialogs, and check them manually. The verification result is shown in the table~\ref{verfication}, based on which we compute the recall, precision and F1 score: 0.970, 0.961, 0.974.

\begin{table}[ht]
\centering
\setlength{\extrarowheight}{0.06cm}
\small
\begin{tabular}[width=\columnwidth]{l|l|l}
\toprule
    & True & False \\
    \hline
 Positive & 97& 3 \\
 \hline
 Negative &96 & 4\\
\bottomrule
\end{tabular}
\caption{Verification of the automated correction of the training/validation set.}
\label{verfication}
\end{table}

\section{Entity Bias}
\label{sec:entity_bias}

As discussed previously, another issue that we observe with MultiWOZ is the entity bias (e.g., ``cambridge" appears in train destination in the majority of dialogs -- Figure~\ref{table_bias}).
Besides the ``train-destination" slot type, we further explore the similar bias problem in all other $30$ slot types. 
For each slot type, we quantify the frequency at which each possible slot value appears in the training data. 

We quantify the entity bias using two metrics. For a vector $r = (r_1, \ldots, r_R)$ of frequencies of $R$ entities, we define the normalized Shannon entropy as
\begin{equation}
    H_1/H_0 :=  \sum_{i \in [R]} r_i \log_{R} \left(\frac{1}{r_i}\right).
\end{equation}
Normalized Shannon entropy is bounded between $0$ and $1$, where $ H_1/H_0 = 0$ implies a deterministic distribution with all weight on a single entity and $1$ implies a perfectly uniform distribution.
Since Shannon entropy does not capture the tail of the distribution, we also report the normalized min-entropy, which is given by
\begin{equation}
    H_{\infty}/H_0 :=  \max_{i \in [R]} \log_{R} \left(\frac{1}{r_i}\right).
\end{equation}
Min-entropy captures the normalized likelihood of the most frequently appearing entity in the list of all possibilities.
For example, as in Figure~\ref{table_bias}, frequency of the entity ``cambridge" is about 50\% which is much higher than 7\% (which would have been its frequency had all $13$ possible entities were uniformly distributed).

The entity bias for all 30 slots in the dataset is depicted in Table~\ref{table_bias}, ordered from the least uniform to the most uniform as measured by {\it normalized Shannon entropy}. 
We observe that some slot types, such as ``hotel parking" and ``hotel-internet" are significantly biased. There are only three possible slot values for the slot type ``hotel-internet": ``yes", ``no" and ``free", where their count numbers are $10,023$, $326$ and $9$ correspondingly. We also find that many other slot types, such as ``restaurant-food" suffer from severe entity bias  besides``train-destination" as well.  
On the other hand, the slot types involving day and people seem to be nicely balanced, such as ``hotel-day" and ``hotel-people". 
This might be because those values are actually uniformly made up, while values of other slot types like ``type", ``food" are real values and indeed follow certain real-world distributions. 
%

\begin{figure}[t]
\centering
\includegraphics[width=0.9\columnwidth]{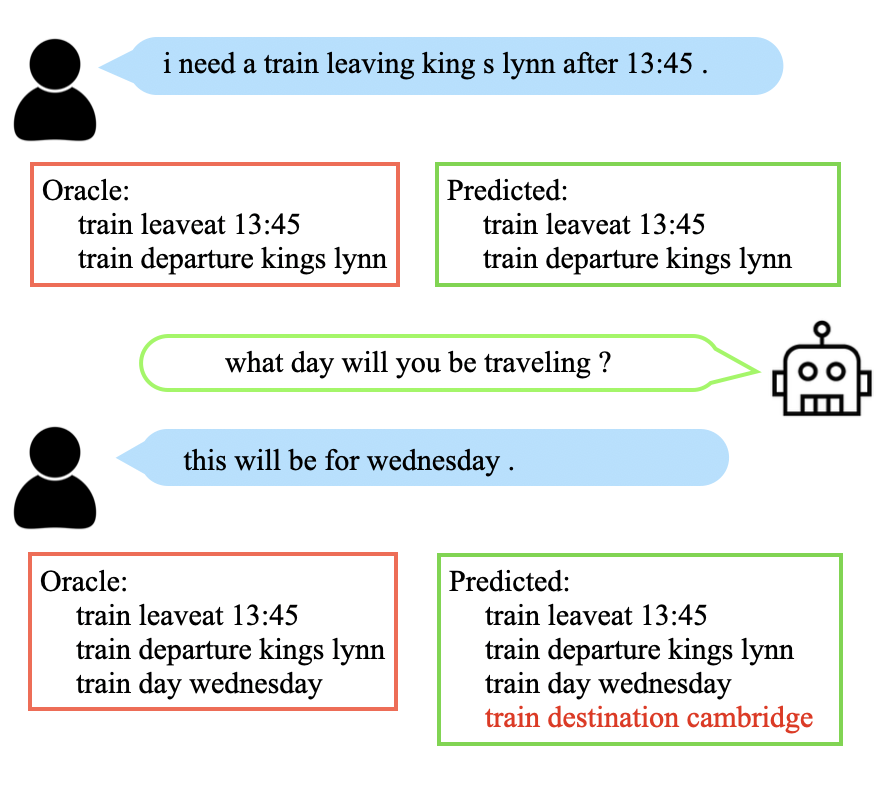}
\caption{\small The first two turns from dialog ``PUML1812" with generated dialog states from the SimpleTOD~\cite{HosseiniAsl2020ASL}, which is trained on MultiWOZ 2.2. The dialog context does not mention ``cambridge", but the model generates  this token. 
}
\label{fig_bias}
\vspace{-.15in}
\end{figure}

These entity biases are potentially amplified by the learning models, which would lead to biased generation.
In Figure~\ref{fig_bias}, we show one such case from SimpleTOD~\cite{HosseiniAsl2020ASL} where in the current turn, the user is providing the information of ``day" in the ``train" domain. 
The dialog state tracking model successfully extracts the token ``wednesday" and updates dialog states in the red rectangle. 
However, the model also adds the dialog state ``train destination cambridge", while ``cambridge" has never been mentioned in the dialog history, which is potentially explained by the severe entity bias present in the ``train-destination."

Different from the annotation inconsistency problem, we do not make any modification to the training dataset based on our observation with respect to the entity bias. Strictly speaking, bias cannot be considered as a source of error in the dataset, and it needs to be tackled via better modeling efforts.
Although the entity bias hurts the prediction accuracy of low-frequency slots and results in generating extra high-frequency slots, it also reflects certain real-world facts/biases as the dialogs are conducted by humans. 
This usually helps the learning task with limited training data, e.g., dialog domain adaptation \cite{Qian2019DomainAD, Lu2020LearningFV}. 

\begin{figure}[t]
\centering
\includegraphics[width=0.9\columnwidth]{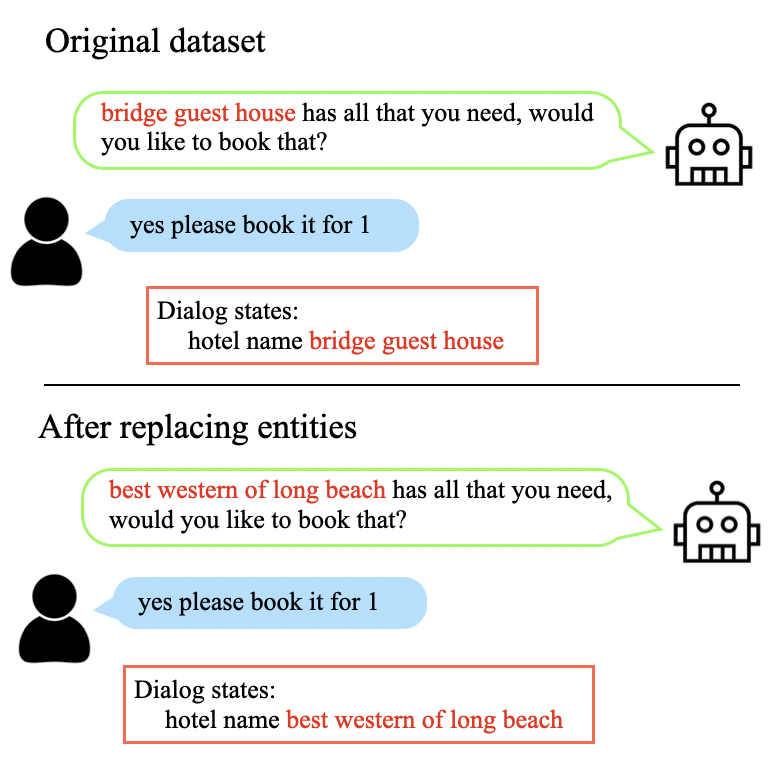}
\caption{\small Example of dialog with new entities by replacing ``bridge guest house" with ``best western of long beach"}
\label{fig_replace_entity}
\vspace{-.15in}
\end{figure}

While we keep the bias in the original dataset intact, we propose a new test set with all entities replaced with new ones unseen in the training data to
facilitate the identification of whether models capitalize on such biases. 
For each slot type from each domain in the MultiWOZ, we find a similar slot type in the Schema-Guided dataset~\cite{rastogi2019towards}. For the slot values belonging to those slot type, we replace them with unseen values from the Schema-Guided dataset. Examples of dialog with replaced entities, along with predicted slots by our benchmark model is shown in Figure~\ref{fig_replace_entity} and Figure~\ref{fig_replace_entity_2} (Appendix).


\begin{table*}[]
\centering
\footnotesize
\scalebox{1.02}{
\begin{tabular}[width=\textwidth]{l|ccc|ccc}
\toprule
\multicolumn{1}{c|}{\multirow{2}{*}{Models}} & \multicolumn{3}{c|}{Standard Results} & \multicolumn{3}{c}{Fuzzy Results}   \\\cline{2-7}
  & 2.1  & 2.2 & Ours  & 2.1 & 2.2 & Ours  \\\hline
TRADE~\cite{WuTradeDST2019}  & 44.4$\pm$0.3          & 45.6$\pm$0.5          & 55.2$\pm$0.2  & 45.1$\pm$0.3        & 46.9$\pm$0.2         & 58.2$\pm$0.4  \\
SimpleTOD~\cite{HosseiniAsl2020ASL}  
& 54.7$\pm$ 0.5 & 53.6$\pm$1.0 & 62.1$\pm$0.2  & 55.2$\pm$ 0.5 & 54.4$\pm$1.2& 64.7$\pm$0.2  \\
DST-BART~\cite{Lewis2020BARTDS}  & 57.9$\pm$0.5         & 56.0$\pm$0.7         & 67.4$\pm$0.5  &  58.7$\pm$0.2           & 57.5$\pm$0.2             & 72.3$\pm$0.4 \\
\bottomrule
\end{tabular}}
\caption{The performance of TRADE, SimpleTOD and DST-BART in terms of joint goal accuracy on MultiWOZ 2.1, 2.2 and our modified version. Fuzzy Results considers model-predicted slot value is correct if it is very similar with the ground truth even if they are not exactly the same.
}
\label{table_benchmark}
\vspace{-.15in}
\end{table*}

\section{Benchmarking State-of-the-Art Models}
\label{sec:experiments}
To verify our corrections of the dialog state annotations, we benchmark state-of-the-art dialog state tracking (DST) models on our modified dataset. 

Traditionally, for DST task, the slot value is predicted by selecting from pre-defined candidates or extracting from dialog context. 
We adopt TRADE~\cite{WuTradeDST2019} as a representative of the mixture of these two methods.
More recent works focus more on fine-tuning pre-trained model, which purely generates slot values based on dialog history. 
We choose SimpleTOD~\cite{HosseiniAsl2020ASL} and fine-tuned BART~\cite{Lewis2020BARTDS} as benchmark models for DST as well.

\noindent\textbf{TRADE}~\cite{WuTradeDST2019} integrates GRU-based~\cite{Cho2014OnTP} encoder-decoder model and pointer-generator~\cite{See2017GetTT} to learn to copy slot values either from the dialog context or from the pre-defined value candidates.

\noindent\textbf{SimpleTOD}~\cite{HosseiniAsl2020ASL}\label{simpletod} builds a DST model by fine-tuning GPT2~\cite{radford2019language}, a large pre-trained language model. It combines all the condition information, including dialog history, previous dialog states and user utterance into a single sequence as input and let the language model learn to generate a sequence, containing dialog states and system response. Here we only feed in dialog states as ground truth output during training step, so that the trained model is specially designed for DST.

\noindent\textbf{DST-BART} builds a DST model by fine-tuning BART~\cite{Lewis2020BARTDS} on the DST task in MultiWOZ. BART is a denoising autoencoder pretrained with corrupted text, making it more robust to noisy data. It consists of a bidirectional encoder and a left-to-right autoregressive decoder.

\vspace{0.03in}\noindent{\bf Joint Goal Accuracy.}
We adopt joint goal accuracy and slot accuracy as our metrics of interest to evaluate the performance of the benchmark models. 
The joint goal accuracy measures the ratio of the dialog turns in the entire test set, where all the slots, in the form of triplets (domain, slot type, slot value) are predicted precisely correctly. 
Instead of checking every dialog turn, slot accuracy checks each slot individually for all slot types. 

The evaluation results are reported in Table~\ref{table_benchmark}. As can be seen, the performance of all baselines increased about 7-10\% where the largest jump happened in TRADE. We also observe that while DST-BART and SimpleTOD are within statistical error on MultiWOZ 2.2, DST-BART outperforms SimpleTOD by 5\% points on our modified dataset. We suspect one potential reason is that BART is pre-trained on corrupted text, which improves its robustness and ability to handle noisy text.
Another 
reason might be that BART consists of an encoder and a decoder, containing 400M parameters, while GPT2 is a decoder only with 117M parameters.

\begin{figure}
\centering
\includegraphics[width=\columnwidth]{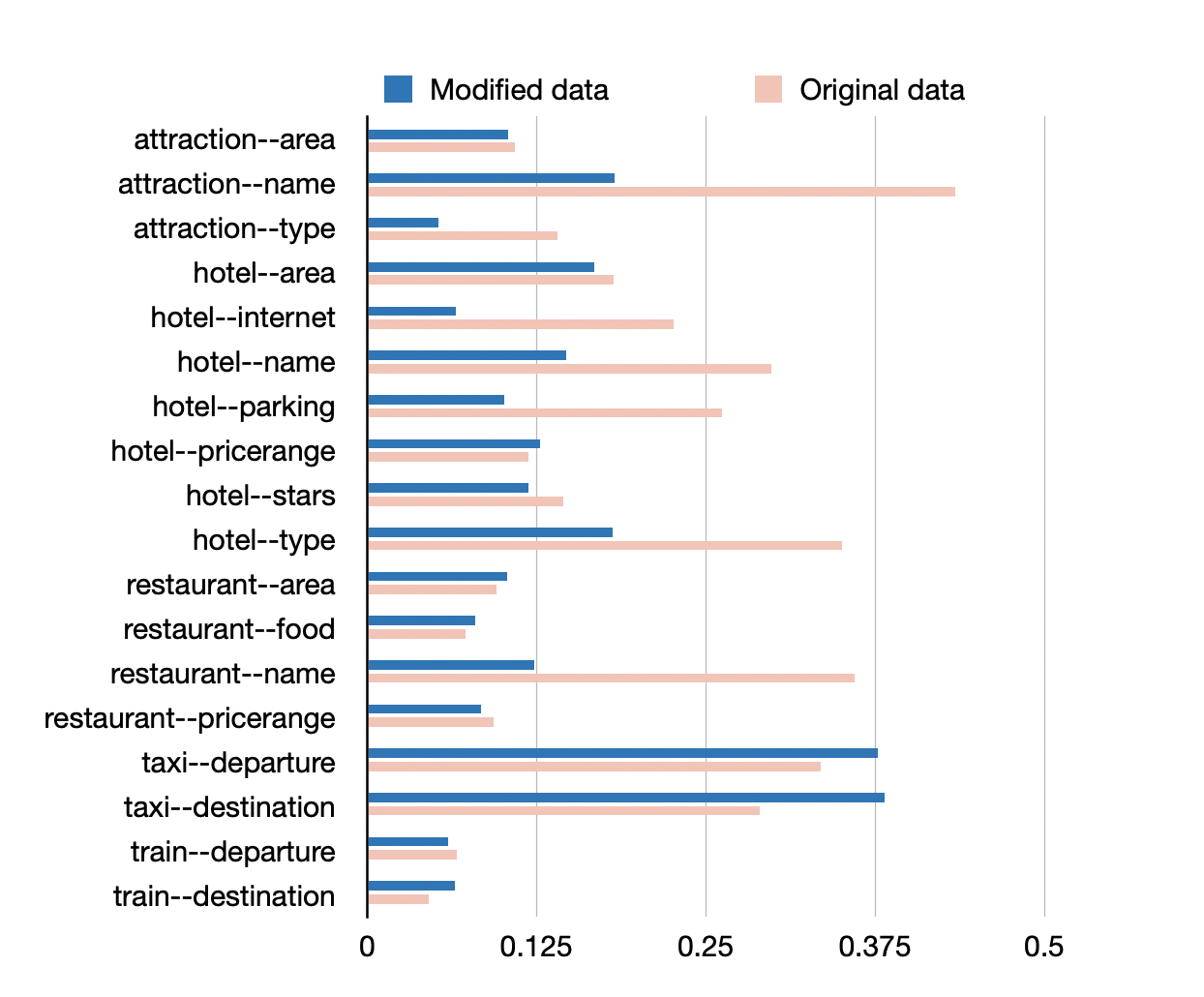}
\caption{Fraction of error turns on the new corrected data.}
\label{fig_ratio_err_turn}
\end{figure}

\begin{figure*}[t]
\centering
\includegraphics[width=\textwidth]{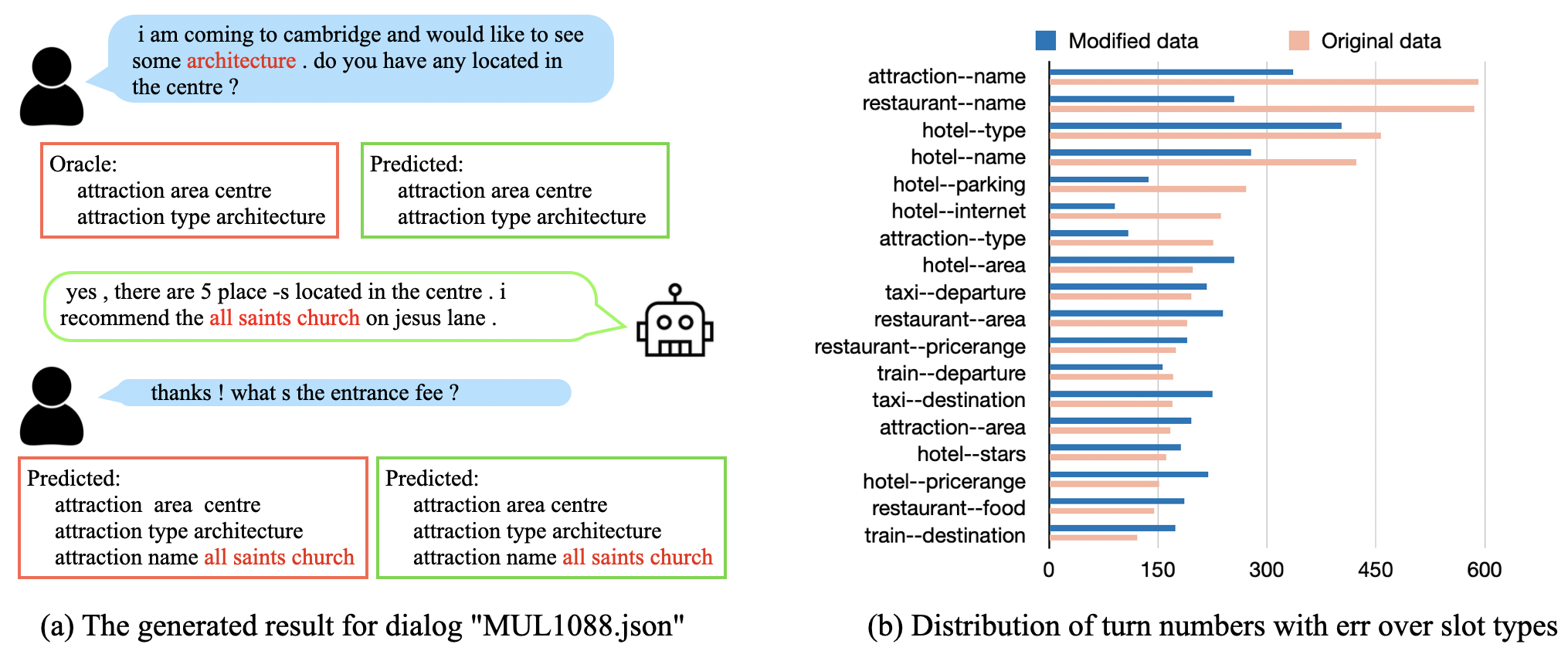}
\caption{\small (a) The generated dialog state results (in green rectangles) using SimpleTOD. The model generates the \textit{``attraction name all saints church"} in the second turn, however, the slot is included in the ground truth after the annotation correction; (b) The distribution of the numbers of turns where SimpleTOD makes a mistake, testing on MultiWOZ 2.2 test set (red bars) and our modified version (blue bars), showing a drop in errors associated with ``attraction-name" and  ``restaurant-name". 
}
\label{fig_after_inc}
\vspace{-.15in}
\end{figure*}

The improvement from previous dataset versions comes mostly from the removal of the confusion because of the inconsistency. With the added slots, some of the predictions previously marked as erroneous are now recognized as correct. As shown in Fig.~\ref{fig_after_inc}(a), the ``attraction name all saints church" was not included in the ground truth in the old dataset version, so the prediction made by SimpleTOD was considered as wrong. With the modified dataset, SimpleTOD makes the same prediction, and since the annotation is more consistent, the models are more confident to learn the pattern and less likely to miss predicting slots. 
The changes in these two aspects leads to the drop of error turn numbers, shown in Fig~\ref{fig_after_inc}(b), especially for the slot type ``name" related, which involves the most slot modifications, corresponding to Table~\ref{table_correct_num}.
The happens at the costs of slight increase of the error turn numbers for other slot types result from the increasing of the total turn number for those slot types, since we add slot annotations in the modified dataset. Overall, the percentages of turns with error all decrease, shown in the Fig.~\ref{fig_ratio_err_turn}.


\vspace{0.03in}\noindent{\bf Fuzzy Match.}
There is another issue that multiple slot values can refer to the same item. For example, in the fifth turn of dialog ``MUL0148.json", the user and system are talking about booking at the ``huntingdon marriott hotel". The ground truth annotation for the hotel name is ``huntingdon marriott", while SimpleTOD predicts ``huntingdon marriott hotel", which is also the value in the dialog context. We believe these kinds of mismatches should be ignored (as they can be fixed via simple wrappers to find the closest match) and attention should be focused to other dominant problems to improve building greatly-performing DST models.
As such, we adopt Levenshtein distance to compute the similarity between the ground truth and the predicted slot values. We consider the prediction to be correct if the similarity is above 90\%. The result is listed in the Table~\ref{table_benchmark}. The performances after the fuzzy matching increases by 1-5\%, consistent with the standard results.

\begin{figure}[t]
\centering
\includegraphics[width=0.9\columnwidth]{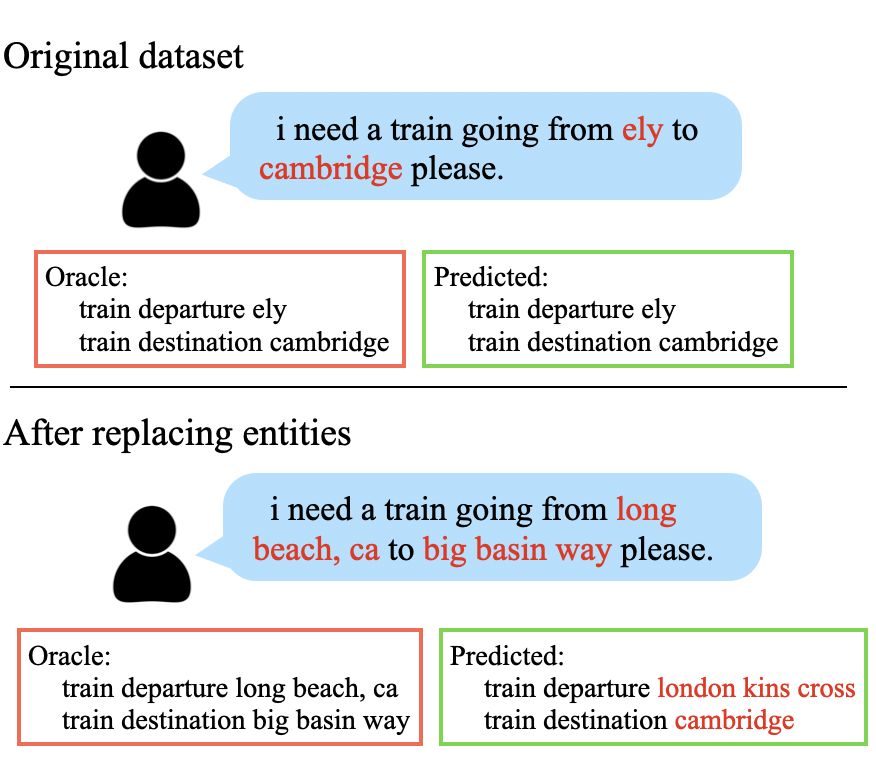}
\vspace{-.1in}
\caption{\small The outputs from DST-BART on the first turn of dialog ``SNG0293.json", before and after replacing the entities.}
\label{fig_err_replace_entity}
\vspace{-.15in}
\end{figure}

\begin{table}[ht]
\centering
\setlength{\extrarowheight}{0.06cm}
\footnotesize
\scalebox{1.02}{
\begin{tabular}[width=\columnwidth]{l|l}
\toprule
     & Joint Goal Acc. \\
    \hline
     MultiWOZ 2.1 test set & 56.0$\pm$0.7 \\
    \hline
 New test set with replaced entities & 27.0$\pm$2.0 \\
\bottomrule
\end{tabular}}
\caption{\small Performance of DST-BART on MultiWOZ 2.1.}
\label{table_replace_entity}
\vspace{-.15in}
\end{table}

\vspace{0.03in}\noindent{\bf New test set.}
Similar with~\cite{raghu-etal-2019-disentangling}, we evaluate SOTA models on a new test set, where we replace the slot entities with unseen values, resulting in a 29\% performance drop in the terms of joint goal accuracy on DST-BART.
In Figure~\ref{fig_err_replace_entity}, we show an example dialog, which was correctly predicted by DST-BART on the original test set. As can be seen, on the new test set, the model predicts entities that never appear in the dialog context hinting at severe memorization of these named entities.

\section{Concluding Remarks}
MultiWOZ is a well-annotated task-oriented dialog dataset, and widely used to evaluate dialog-related tasks. Previous works like MultiWOZ 2.1 and MultiWOZ 2.2 have carefully identified and corrected several errors in the dataset, especially for the dialog state annotations. Building on MultiWOZ 2.2, we identified annotation inconsistency across different dialogs as a source of confusion for training dialog state tracking models. We proposed a correction and released a new version of the data with corrections. We also identified named entity bias as another source of issue, and released a new test set with all named entities replaced with unseen ones. Finally, we benchmarked a few state-of-the-art dialog state tracking models on the new versions of the data, showing 5-10\% performance improvement on the new corrected data, and 29\% performance drop when evaluation is done on the new test set with replaced entities. 
We hope the better understanding of MultiWOZ helps us gain more insights into dialog evaluation on this dataset.

While we corrected some errors in this work, we observe a few remaining problems in MultiWOZ.
First, there are some cases where the annotation contradicts with the database. For example, in  dialog ``MUL2523.json", the user is asking about ``autumn house", which is annotated as type of ``guest house" in the database. However, the dialog state annotation labels the hotel type as ``hotel". This disagreement might not hurt the training of dialog state tracking models, but affects the now popular end-to-end dialog models, which are trained on MultiWOZ.
There still some annotation errors. For example, in dialog ``MUL0072.json", the ``monday" has never been mentioned, while the dialog states include the slot ``hotel day monday".

Finally, the source of the inconsistencies identified in this work is the Wizard of Oz data collection strategy, where different crowd-workers may annotate the dialogs differently. One way to mitigate such confusions might be to provide annotators with crystal clear annotation guidelines, or to have each dialog annotated by multiple annotators.

    

\bibliographystyle{acl_natbib}
\bibliography{anthology,acl2021}

\newpage
~
\newpage
\appendix
\onecolumn


\begin{figure*}[t]
\centering
\includegraphics[width=0.9\textwidth]{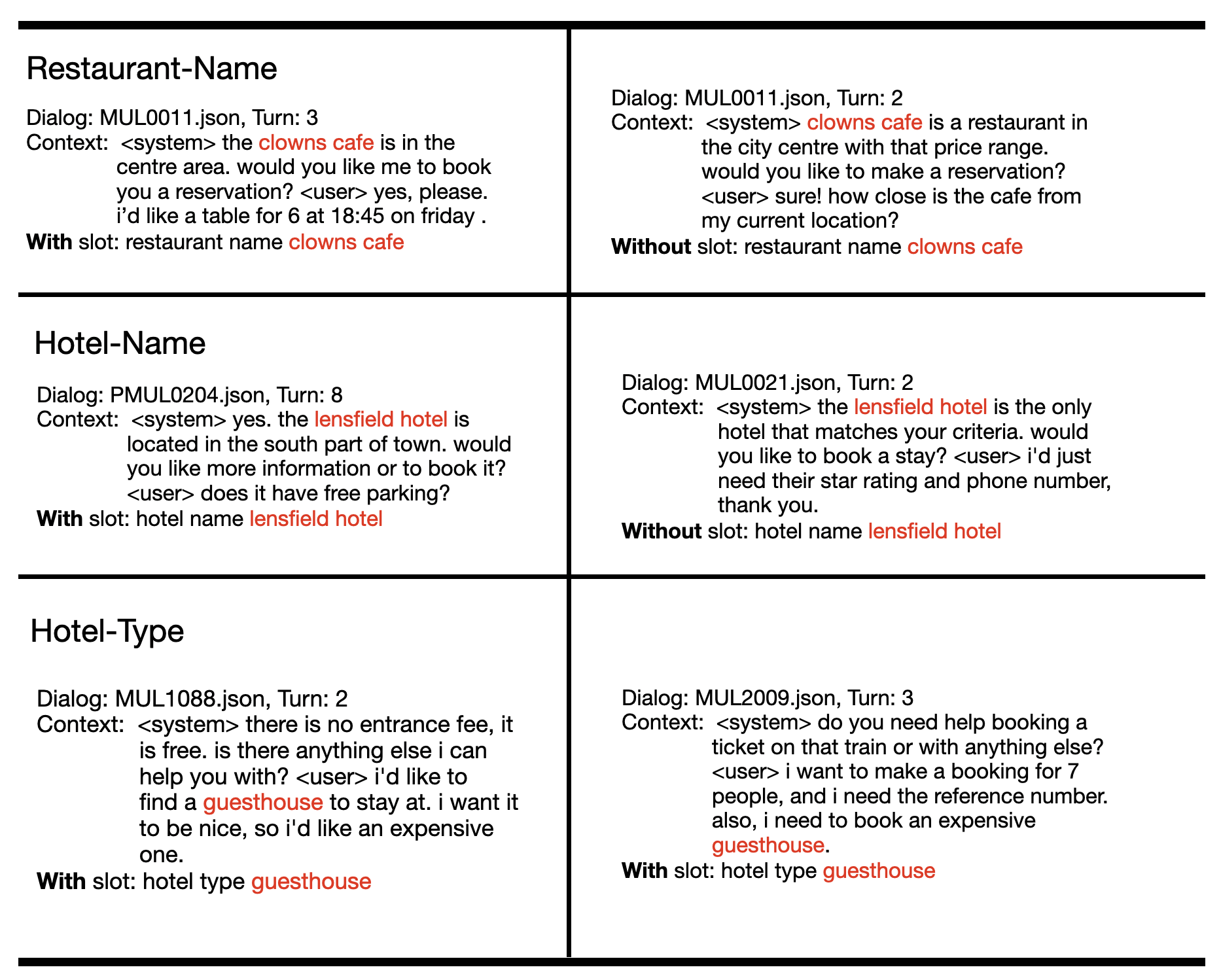}
\caption{More examples of dialog with or without slot annotations for different slot types.}
\label{fig_more_examplt}
\end{figure*}


\begin{table*}[h]
\centering
\setlength{\extrarowheight}{0.06cm}
\small
\begin{tabular}[width=\textwidth]{c|l|lll}
\toprule
domain                      & slot type   & train           & valid          & test           \\\hline
\multirow{4}{*}{attraction} & area        & 2081 (13.18\%)  & 367 (15.17\%)  & 350 (14.33\%)  \\
                            & name        & 3738 (23.67\%)  & 558 (23.06\%)  & 505 (20.67\%)  \\
                            & type        & 2915 (18.46\%)  & 426 (17.60\%)  & 468 (19.16\%)  \\\cline{2-5}
                            & total       & 7288 (46.15\%)  & 1163 (48.06\%) & 1078 (44.13\%) \\\hline
\multirow{8}{*}{hotel}      & area        & 3486 (15.72\%)  & 530 (18.96\%)  & 454 (17.37\%)  \\
                            & internet    & 3175 (14.32\%)  & 400 (14.31\%)  & 340 (13.01\%)  \\
                            & name        & 3998 (18.03\%)  & 616 (22.03\%)  & 479 (18.32\%)  \\
                            & parking     & 3122 (14.08\%)  & 431 (15.41\%)  & 326 (12.47\%)  \\
                            & pricerange  & 4342 (19.59\%)  & 485 (17.35\%)  & 443 (16.95\%)  \\
                            & stars       & 2989 (13.48\%)  & 472 (16.88\%)  & 406 (15.53\%)  \\
                            & type        & 7061 (31.85\%)  & 985 (35.23\%)  & 919 (35.16\%)  \\\cline{2-5}
                            & total       & 15048 (67.88\%) & 1929 (68.99\%) & 1822 (69.70\%) \\\hline
\multirow{5}{*}{restaurant} & area        & 2906 (12.57\%)  & 435 (14.99\%)  & 352 (12.21\%)  \\
                            & food        & 2648 (11.45\%)  & 361 (12.44\%)  & 349 (12.11\%)  \\
                            & name        & 4705 (20.35\%)  & 702 (24.20\%)  & 458 (15.89\%)  \\
                            & pricerange  & 3036 (13.13\%)  & 441 (15.20\%)  & 367 (12.73\%)  \\\cline{2-5}
                            & total       & 9932 (42.95\%)  & 1363 (46.98\%) & 1051 (36.46\%) \\\hline
\multirow{3}{*}{taxi}       & departure   & 32 (0.70\%)     & 2 (0.29\%)     & 0 (0.00\%)     \\
                            & destination & 23 (0.50\%)     & 1 (0.15\%)     & 1 (0.16\%)     \\\cline{2-5}
                            & total       & 53 (1.16\%)     & 3 (0.44\%)     & 1 (0.16\%)     \\\hline
\multirow{3}{*}{train}      & departure   & 235 (1.29\%)    & 14 (0.48\%)    & 35 (1.19\%)    \\
                            & destination & 441 (2.43\%)    & 61 (2.10\%)    & 5 (0.17\%)     \\\cline{2-5}
                            & total       & 658 (3.62\%)    & 75 (2.59\%)    & 35 (1.19\%)    \\\hline
\multicolumn{2}{c|}{total}                 & 27656 (50.50\%) & 3783 (51.92\%) & 3467 (47.68\%)\\

\bottomrule
\end{tabular}
\caption{Number of Turns involving modification for inconsistency distribution.}
\label{table_correct_dial_turn_num}
\end{table*}

\begin{figure*}[t]
\centering
\includegraphics[width=\textwidth]{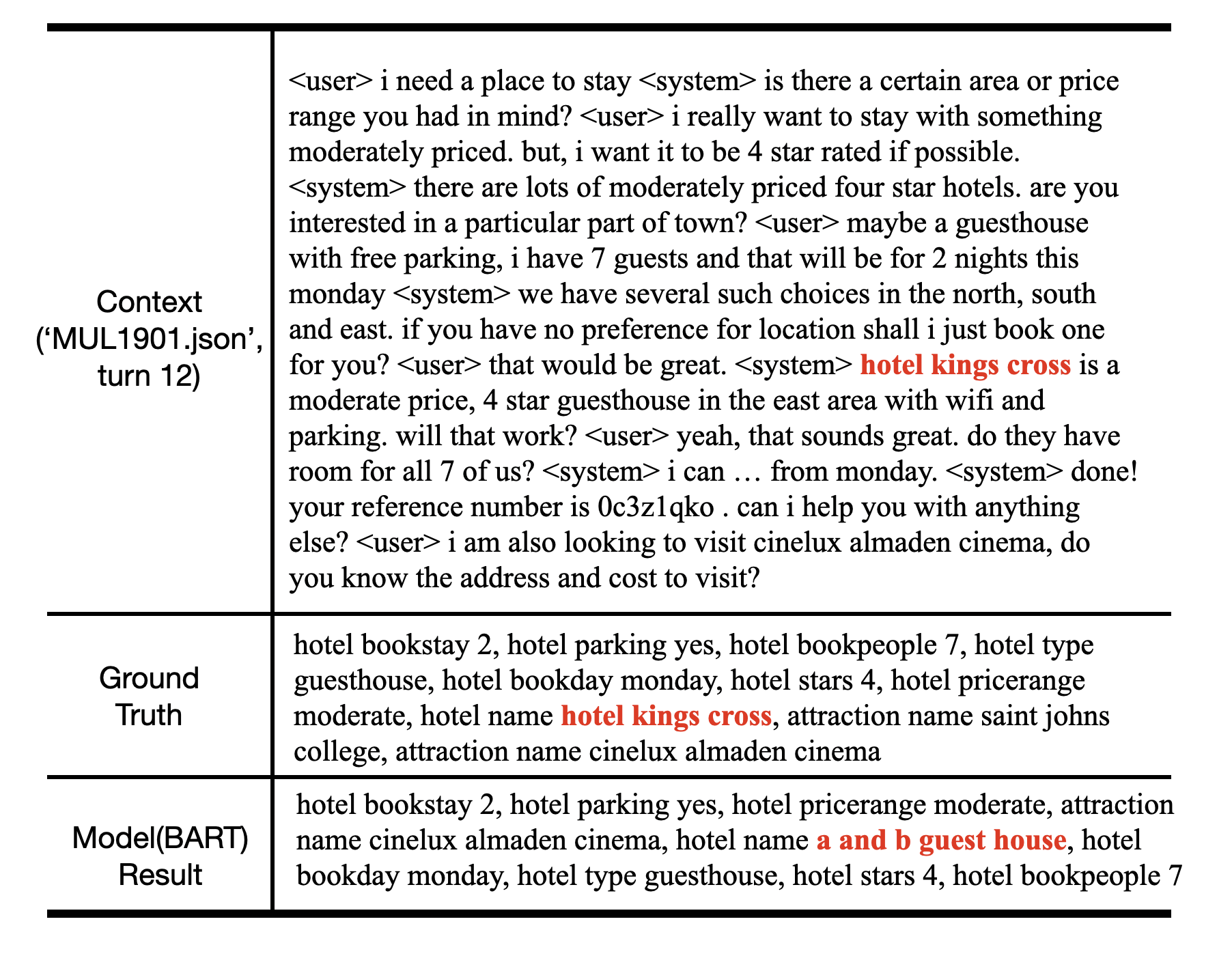}
\caption{Example of dialogs with new entities}
\label{fig_replace_entity_2}
\end{figure*}

\label{sec:appendix}





\end{document}